\documentclass[a4paper,11pt,twocolumn,twoside]{article}
\usepackage{graphicx}
\usepackage{sepln_en}
\usepackage{fullname_esp}
\usepackage[utf8]{inputenc}
\usepackage[T1]{fontenc}
\usepackage{rotating}
\usepackage{makecell}
\usepackage{caption}
\usepackage{hyperref}

\DeclareUnicodeCharacter{0301}{\'{e}}

\input epsf

\title{Deep Learning Methods for Extracting Metaphorical Names of Flowers and Plants}

\author {\textbf{Amal Haddad Haddad$^1$,} \textbf{Damith Premasiri$^2$,} \textbf{Tharindu Ranasinghe$^3$,} \textbf{Ruslan Mitkov$^4$}\\
$^1$University of Granada, Spain\\
$^2$University of Wolverhampton, UK\\
$^3$Aston University, UK\\
$^4$Lancaster University, UK\\
amalhaddad@ugr.es\\
}

\seplntranstitle{Métodos de aprendizaje profundo para la extracción de nombres metafóricos de flores y plantas}

\seplnclave{Aprendizaje profundo, Transformadores, Extracción automática de metáfora, Términos basados en metáfora}

\seplnresumen{El dominio de la Botánica es rico en términos metaforicos. Estos términos tienen un papel importante en la descripción e identificación de flores y plantas. Sin embargo, la identificación de este tipo de términos en el discurso es una tarea difícil. Esto puede conducir a errores en los procesos de traducción y otras tareas lexicográficas.  Este proceso es aún más difícil cuando se trata de traducción automática, tanto en el caso de las unidades monoléxicas, como en el caso de las unidades multiléxicas. 
Uno de los desafíos a los que se enfrentan las aplicaciones del Procesamiento del Lenguaje Natural y las tecnologías de Traducción Automática es la identificación de términos basados en metáfora a través de métodos de aprendizaje profundo. En este estudio, tenemos el objetivo de rellenar este vacío a través del uso de trece modelos populares basados en transformadores, además del ChatGPT. Asimismo, demostramos que los modelos discriminativos aportan mejores resultados que los modelos de GPT-3.5. El mejor resultado alcanzó una puntuación de 92,2349\% F1 en las tareas de identificación de nombres metafóricos de flores y plantas.}

\seplnkey{Deep Learning, Transformers, Automatic Extraction of Metaphor, Metaphor-based Terms}

\seplnabstract{The domain of Botany is rich with metaphorical terms. Those terms play an important role in the description and identification of flowers and plants. However, the identification of such terms in discourse is an arduous task. This leads in some cases to committing errors during translation processes and lexicographic tasks. The process is even more challenging when it comes to machine translation, both in the cases of single-word terms and
multi-word terms. One of the recent concerns of Natural Language Processing (NLP) applications and Machine Translation (MT) technologies is the automatic identification of metaphor-based words in discourse through Deep Learning (DL). In this study, we seek to fill this gap through the use of thirteen popular transformer based models, as well as ChatGPT, and we show that discriminative models perform better than GPT-3.5 model with our best performer reporting 92.2349\% F1 score in metaphoric flower and plant names identification task.}

\firstpageno{1}

\begin{document}


\setlength\titlebox{21cm} 

\label{firstpage} \maketitle

%

\section{Introduction}
Metaphor is a pervasive phenomenon in human language \cite{lakoff2008metaphors}. It is defined as "mapping of conceptual structure from a source to a target domain" \cite{ruiz2017conceptual}. Depending on the dimension of complexity of metaphor, authors distinguish two types of metaphors: image metaphors and conceptual metaphors \cite{lakoff2008metaphors}. Image metaphors compare one single image in one domain with another image belonging to another domain, such as the image metaphor "she is as good as gold". Conceptual metaphors are more complex at the conceptual and cognitive levels, and they refer to the resemblance established between a whole set of experiences, such as the metaphor in "life is a journey", which implies a whole set of elements activated within the metaphoric target domain. 
Metaphor-based terms, or the so-called terminological metaphors, are common in specialised languages. Their use is abundant, as they help in the conceptualisation of phenomena and their description by establishing a resemblance between images and domains. They also help in understanding abstract phenomena in terms of more concrete notions and in modelling scientific thought \cite{urena2010strategies}. However, the identification of metaphor-based terms in discourse is an arduous task. This leads in some cases to committing errors during translation processes and lexicographic tasks. The process is even more challenging when it comes to machine translation, both in the cases of single-word terms and multi-word terms, which are represented by Multiword Expressions (MWEs). The main common error while carrying out the translation processes is that the metaphorical lexical items forming part of a term would be transferred literally into other languages without taking into consideration its metaphoric and cultural dimension or without taking into account that they form part of an MWE.

Previous studies focused on the extraction of metaphorical terms from discourse, such as \namecite{mu-etal-2019-learning} and \namecite{razali2022deep}; however, to the best of our knowledge, there are no programs that could automatically retrieve those terms both as single-word terms and MWEs in specialised languages. This study seeks to fill in this gap and proposes a novel method based on transformer models \cite{premasiri2022mumtt,premasiri2022bert}; \cite{ranasinghe-etal-2021-wlv} for automatic extraction of metaphor-based terms from the specialised domain of Botany and concerning the names of flowers and plants in English and Spanish.
The main contributions of this study are: 
\begin{enumerate}
    \item We empirically evaluate thirteen discriminative transformer models and one generative transformer model (ChatGPT) for the tasks of metaphoric flower and plant names identification on English and Spanish datasets. 
    \item We show that discriminative models perform better in the metaphoric flower and plant names identification task.
    \item We release new annotated datasets for metaphoric names identification in English and Spanish. 
    \item We make our code freely available for further research\footnote{\url{https://bit.ly/3pYAYXK}}.
\end{enumerate}

This paper is organised as follows: in Section \ref{sec:related} we present previous related work. In Section \ref{sec:data} we describe the dataset used and its annotation process. In Section \ref{sec:methods} we detail the experimental set-up and methodology, while in Section \ref{sec:results} we report our experiment’s results and evaluation. Finally, we summarise the main conclusions and propose future work in Section \ref{sec:conclusions}.

\section{Related work}
\label{sec:related}
The study of metaphor-based terms in discourse has been a subject of study in the last few decades. One of the main concerns in this field is the detection of metaphor-based words in discourse. With this aim, the Pragglejaz Group suggested a method for the manual identification of metaphor, called Metaphor Identification Procedure (MIP) \cite{group2007mip}. This method has been used extensively \cite{nacey2019metaphor}. Studies like \namecite{turney2011literal}, \namecite{jang2015metaphor} and \namecite{coll2019new} have a similar approach. Other projects such as the VU Amsterdam Metaphor Corpus \cite{leong2020report} offer a manually annotated corpus for all metaphorical language use. Moreover, studies like \namecite{yaneva2016assessing}, show how the use of metaphor and figurative language in discourse is of utmost difficulty for people with Autism Spectrum Disorder (ASD); hence, studies like \namecite{yaneva2016assessing} and \namecite{vstajner2017effects} endeavour to identify and disambiguate complex sentences which contain metaphor and metonymy among other features through the application of Complex Word Identification modules. The above studies were partially inspired by the FIRST Project\footnote{\url{http://www.iwebtech.co.uk/project-first/}} \cite{oruasan2018intelligent} and the development of the Open Book tool which helps people with ASD.

Concurrently, one of the recent concerns of Natural Language Processing (NLP) applications and Machine Translation (MT) technologies is the automatic identification of metaphor-based words in discourse through Deep Learning Methods (DLM). For example, \namecite{mu-etal-2019-learning} suggest working with large corpora and training \textit{simple gradient boosting classifiers} on representations of an utterance and its surrounding discourse learned with a variety of document embedding methods”. \namecite{su2020deepmet} focus on token-level metaphor detection paradigm and propose using an end-to-end deep metaphor detection model. Authors like \namecite{razali2022deep} use machine learning to automatically detect metaphor instances in short texts by implementing Support Vector Machine algorithms, while other authors like \namecite{gutierrez2016literal} propose modelling metaphor explicitly within compositional distributional semantic models to improve the resulting vector representations. Those authors classify the already used methods in the following categories: clustering; topic modelling; topical structure and imageability analysis; semantic similarity graphs and feature-based classifiers \cite{gutierrez2016literal}. Recent approaches are more centred on using dense embedding methods \cite{vitez2022extracting}.

On the other hand, the study of metaphor-based terms in specialised discourse has been subject to scientific and cognitive studies. The automatic identification of metaphor-based terms is considered a substantial challenge. Some studies highlight the importance of automatic extraction of terms in specialised discourse \cite{rodriguez2005metalinguistic} while other studies, such as \namecite{urena2010strategies}, propose a semi-automatic method for term retrieval in the domain of Marine Biology. However, to the best of our knowledge, there have been no previous studies or methodologies which cover the automatic extraction of those terms from scientific discourse in other domains and no previous studies were carried out in the domain of Botany.

\section{Data}
\label{sec:data}
Specialised discourse is rich in metaphor-based terms; Botany is no exception. The semantic motivations for plant names are usually influenced by the appearance of the plant, the place of its occurrence, the properties of the plant, its usage, as well as other motivations typical of a specific genus of species \cite{dkebowiak2019semantic}. Many studies have shown that metaphor is one of the most frequent techniques to coin flowers and plants names \cite{rastall1996metaphor}; \cite{nissan2014multilingual}; \cite{dkebowiak2019semantic}. This metaphoric use may give clues to cultural references related to legends and beliefs associated with plants in general, like their healing properties and supposed magical powers \cite{dkebowiak2019semantic}. At the same time, this shows that this metaphorical use may vary among languages and cultures. From another perspective, studies like \namecite{goodman1963malayalam} highlight the importance of flower names based on metaphor for the study of colour and its comparison among languages. For this reason, we consider the study of metaphor-based terms in this domain relevant as a case-study.

The dataset we use to extract metaphor-based terms in English is the Encyclopaedia of Flowers and Plants, published by the American Horticultural Society \cite{flowerencyclopedia}. We selected this edition as it is available in a digitalised format in the online library of the Internet Archive. This Encyclopaedia consists of 522,707 words. It contains a dictionary of names of flowers from around the world, with approximately 8000 terms referring to both scientific and common names and their origins, as well as 4000 images. It is divided into the following sections: firstly it has an introduction about how to use the book, plant names and origins and relevant information on how to create a garden and how to select plants. This introductory part shows that it is aimed at both professionals and laypersons. Secondly, it has a plant catalogue, subdivided into categories such as trees, shrubs, roses, climbers and wall shrubs, perennials, annuals, biennials and bedding, rock plants, bulbs, water and bog plants as well as tender and exotic plants. All those subsections contain rich contexts on each term, concerning the origin, uses, habitat, size, etc. Finally, the Encyclopaedia offers a dictionary section with an index of common names and glossary of terms. We benefited from this last section to extract and annotate terms. The advantage of using this Encyclopaedia is that it includes a wide range of varieties of flowers and plants from all around the world. For this reason, the obtained results may be useful to be applied in different contexts and in multidisciplinary studies.

The data was pre-processed by annotating the proper names and their metaphorical condition. The MIP criteria for metaphor identification \cite{group2007mip} was adapted to annotate the terms, considering a term as metaphor-based when one or more of the lexical units forming it or its etymology give evidence that they belong to different domains, based on its meaning in the dictionary. The annotated names represent both image metaphors and conceptual metaphors. An example of image metaphors, is the one-word name of the flower \textit{Edelwiess} which is a combination between the two lexical units \textit{edel} which means noble and \textit{weiss}, which means white in German. This name represents an image metaphor where the flower is called as so as it symbolises purity. The scientific name of this flower is \textit{Leontopodium Alpinum}, an MWE with Greek origin and etymology. It is also  an image metaphor, as the lexical unit \textit{Leontopodium} means lion’s foot \cite{dweck2004review}, the resemblance is established between the for of the petals of the flowers and the aspect of the foot of a lion. Another example are the flowers \textit{Sunburst} and \textit{Moonlight}. The name of the flower \textit{Sunburst} shows the resemblance between the colours of the flower and the colours of the sun, while the flower called \textit{Moonlight}, alludes to the resemblance between the flower and the light of the moon.
Other metaphor based-names represent a conceptual image, such as the MWE flower name \textit{forget-me-not} which refer to the association between the heart-shaped blue flowers that reminds the person of his or her beloved one; or the one-word name of the flower \textit{cascade} which associate the aspect of a flower with the whole process of the water falling in a real cascade.

Apart from the Encyclopaedia of Plants and Flowers, we also compiled a corpus of other resources related to Botany in English. It consists of 437,663 words. Some of the texts are monographs, others are journal articles, and some texts are retrieved from other online resources. The full list of references used to compile the English corpus are listed in Appendix 1.
With respect to the Spanish dataset, we have annotated a list of flowers and plants names provided in selected monographs and glossaries following the same criteria as in the case of the English terms. Above all, we used books and articles in the domain of Botany and botanical glossaries, such as the glossaries provided in \textit{Los Áraboles en España} \cite{Arboles}, \textit{Biología de la Conservación de Plantas en Sierra Nevada} \cite{penas2019biologia} and the glossary of scientific names of plants and their vernacular names provided by the Entomological Museum in Leon in the Bio-Nica webpage\footnote{\url{http://www.bio-nica.info/home/index.html}}. The list obtained from this source consists of more than 5000 scientific and vernacular names of flowers and plants. 
As for the book \textit{Los Áraboles en España}, it consists of almost 155,000 words with more than 600 terms in the section of Glossary. The book describes the details of each plant, its family names, its vernacular names and synonyms, its origin, etymology, description and cultivation information. It also provides illustrative images of each plant. The book \textit{Biología de la Conservación de Plantas en Sierra Nevada} was also valuable as some of its chapters contained lists of scientific names of endemic flowers from Sierra Nevada and its common names too. In order to enhance the datasets, we also added more specialised, semi-specialised and informative texts in the domain of botany to obtain more rich contexts. It consists of 460,258 words. The full list of the sources used to compile the Spanish corpus are listed in Appendix.

With this paper, we release datasets of English and Spanish flower and plant names with their annotations metaphoric or not metaphoric. The English dataset consists of 6330 total plant and flower names as a combination of 1869 metaphorical names and 4461 non-metaphorical names. The Spanish dataset consists of 875 metaphoric names and 4,988 non-metaphoric names out of 5863 total.

\paragraph{Data Preparation}
Since we model the metaphoric name identification task as a token level classification task, we used IOB format tagging for our corpus. IOB format is widely used in token level classification tasks \cite{tjong-kim-sang-de-meulder-2003-introduction} where B - Beginning, I - Inside and O - outside of a metaphoric flower or plant name; Table \ref{tab:annotation_sample} shows an example IOB annotation.
After tagging the sentences from the corpus, we identified that there were a very high number of sentences which do not have a single metaphoric name. In other words, the majority of the sentences only had 'O' as the tag for all their words. Since this has a negative impact on the model training process, we decided to balance the dataset by removing some sentences. Then we shuffled all the sentences and divided the training and test sets. Finally, we had 2020 total sentences divided 1500 and 520 in English training and test set respectively. For Spanish, we used only 250 sentences as the dataset.

\begin{table*}[ht]
    \centering
    \begin{tabular}{c|c|c|c|c|c|c|c|c|c}
    \hline
         calliandra & haematocephala & (Red & powder & puff) & is & an & evergreen, & spreading & shrub   \\
         \hline
         O & O & B & I & I & O & O & O & O & O\\
         \hline
    \end{tabular}
    \caption{BIO annotation example}
    \label{tab:annotation_sample}
\end{table*}

Test sets were the same for discriminative and generative experiments. The only thing is that in the generative approach, we did not use the training set, since we cannot train ChatGPT.

\section{Methodology}
\label{sec:methods}
\begin{figure}[ht]
    \centering
    \includegraphics[scale=0.50]{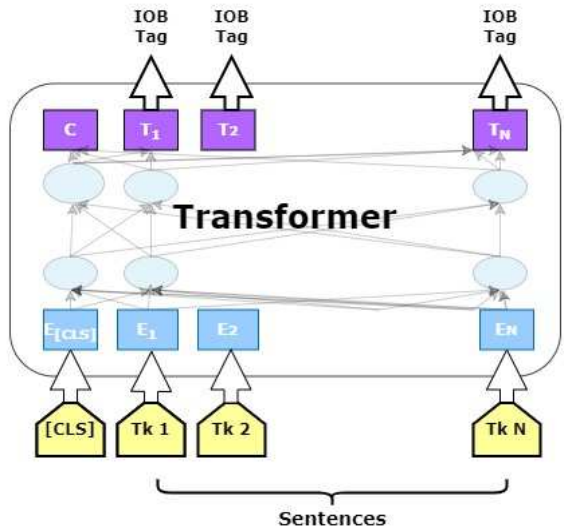}    \caption{Transformers architecture for token level classification}
    \label{fig:transformer}
\end{figure}

\paragraph{\protect Discriminative Models}
Transformers \cite{vaswani2017attention} have been a major breakthrough in Deep Learning research, since they provide a robust mechanism based on attention for the neural networks to flow information without recurrence and convolution. This architecture has produced state-of-the-art results in many NLP applications. With the introduction of BERT \cite{devlin-etal-2019-bert}, which employs the transformers architecture, the pre-trained large language models have played an important role in pushing the boundaries of all NLP tasks such as text classification \cite{ranasinghe2019brums} , \cite{uyangodage-etal-2021-multilingual},  question answering \cite{premasiri-etal-2022-dtw}, text similarity \cite{mitkovautomatic} etc. and achieving new state-of-the-art. With this motivation, we use transformers as our primary experimental setup and evaluate multiple pre-trained language models. These models follow similar architectures to BERT \cite{devlin-etal-2019-bert} while they are pre-trained on different corpora and different objectives. Figure \ref{fig:transformer} \cite{ranasinghe-zampieri-2021-mudes} shows the transformer architecture we used where we input sentences which contain metaphoric flower and plant names, then we obtain BIO tags from the output layer by adding a softmax layer on top of the last hidden state of the deep network to classify each token into one of I,O,B tags. We used several popular transformers based pre-trained language models.

For the experiments on English dataset, we used the cased and uncased variants of BERT base and BERT large versions. In order to establish the capabilities of multilingual models, we experimented with the multilingual-bert \cite{devlin-etal-2019-bert} model with its cased and uncased variants and xlm-roberta-base \cite{conneau-etal-2020-unsupervised} model and xlm-roberta-large \cite{conneau-etal-2020-unsupervised} version. We further experimented with google/electra-base-discriminator \cite{clark2020electra} model which is different from BERT architecture. Finally, within these discriminative models we evaluate allenai/scibert\_scivocab\_cased \cite{beltagy-etal-2019-scibert} and allenai/scibert\_scivocab\_uncased \cite{beltagy-etal-2019-scibert} variants which are specifically pre-trained on scientific corpora. We assume that flower and plant names could appear in those corpora such that the model can leverage the learning to produce better results. 

Since Spanish is low in resources on metaphoric flower and plants names corpora, we experimented zero-shot learning for Spanish on English data. We specifically used the multilingual-bert \cite{devlin-etal-2019-bert} and xlm-roberta \cite{conneau-etal-2020-unsupervised} for our experimental setting as these models provide multilingual capabilities. 

All the models were trained for three epochs, learning rate 4e-5 with 32 training batch size and for the hardware we used a GeForce RTX 3090 GPU.

\begin{table*}[t]
    \centering
    \begin{tabular}{c|c|c|c}
        \hline
         Model & Precision & Recall & F1 \\
         \hline
         bert-base-uncased & 92.8204 & 89.4824 & 91.0784 \\
         bert-base-cased & 93.4157 & 90.8295 & 92.0801 \\
         bert-large-uncased & 92.8424 & 90.6789 & 91.7219 \\
         bert-large-cased & 93.4157 & 90.8295 & 92.0801 \\
         bert-base-multilingual-uncased & 91.7655 & 89.6286 & 90.6648 \\
         bert-base-multilingual-cased & 93.3662 & 91.1718 & \textbf{92.2349} \\
         xlm-roberta-base & 90.1220 & 89.6020 & 89.8560 \\
         xlm-roberta-large & 90.8455 & 89.4348 & 90.1220 \\
         xlnet-base-cased & 89.8189 & 90.8769 & 90.3402 \\
         roberta-base & 91.9779 & 89.8922 & 90.9025 \\
         google/electra-base-discriminator & 92.0412 & 91.1617 & 91.5898 \\
         allenai/scibert\_scivocab\_uncased & 91.7084 & 90.3453 & 91.0071 \\
         allenai/scibert\_scivocab\_cased & 92.3408 & 90.6466 & 91.4750 \\
         \hline
         ChatGPT & 62.1516 & 45.1943 & 48.1392 \\
         \hline
    \end{tabular}
    \caption{Resutls for English metaphoric flower and plant names identification; the Model column represents the model we experimented, the Precision column shows the macro precision, the Recall column shows macro recall and the F1 column shows macro F1 value for the results}
    \label{tab:english_results_table}
\end{table*}

\begin{table*}[t]
    \centering
    \begin{tabular}{c|c|c|c}
          \hline
        Model & P & R & F1 \\
        \hline
        bert-base-multilingual-uncased & 59.2957 & 40.3103 & 43.0472\\
        bert-base-multilingual-cased & 54.0904 & 52.1401 & \textbf{52.8657} \\
        xlm-roberta-base & 67.4035 & 36.5622 & 37.4988 \\
        xlm-roberta-large & 64.1040 & 47.4813 & 51.8174 \\
        \hline
        ChatGPT & 63.1887 & 46.6820 & 51.4120 \\
        \hline
    \end{tabular}
    \caption{Results on metaphoric flower and plant names identification in Spanish; P - The macro averaged precision, R - The macro averaged Recall, F1 - The macro averaged F1 score.}
    \label{tab:spanish_results}
\end{table*}

\paragraph{Generative Models}
While all above methods rely on the discriminative approach, which tries to identify boundaries in the data space, generative models attempt to model the placement of the data throughout the space. This approach attracted huge attention in the research community with the release of ChatGPT\footnote{\url{https://chat.openai.com/}} by openAI\footnote{\url{https://openai.com/}}. The research on Generative Pre-trained Transformer (GPT) \cite{radford2018improving} models have produced multiple versions of it including GPT-3, GPT-3.5 and GPT-4. The free version of ChatGPT only supports GPT-3.5 for the time being and all our experiments are based on ChatGPT free version. 
According to OpenAI, the most cost capable and cost effective models out their models is gpt-3.5-turbo, which we used to our experiments. 

Since ChatGPT is a generalised conversational application, it does not essentially provide IOB tags as outputs. After experimenting with different prompts to retrieve IOB tags from ChatGPT, we decided it would be easier to retrieve the metaphoric flower or plant name in the sentence from the API\footnote{\url{https://bit.ly/3OLCWFn}} and \textit{No} otherwise. Prompt we used: \textit{Is there a metaphoric flower name or metaphoric plant name included in the following sentence, say yes or no, if yes what is the metaphoric flower or metaphoric plant names in the sentence separately : \{sentence goes here\}}. The outputs of ChatGPT are not uniform, and we had to post process the outputs using regular expressions to re-generate the IOB tags for evaluation. 

Since this is a token classification task, we use macro averaged Precision, Recall and F1 score as our evaluation metrics.

\begin{equation}
Precision = TP/(TP + FP)
\end{equation}
\begin{equation}
Recall = TP/(TP + FN)
\end{equation}
\begin{equation}
\resizebox{0.95\hsize}{!}{
F1 = 2 * (Precision * Recall)/(Precision + Recall)
}
\end{equation}

\section{Results and Discussion}
\label{sec:results}

\subsection{English}
The results in table \ref{tab:english_results_table} show the competitive performance of transformer models, in the flower and plant names classification task. Despite the fact that most of the transformer models we experimented with are not specifically pre-trained on botanic corpora, almost all discriminative models were able to produce more than 90\% F1 score in the task. Interestingly, the multilingual bert model could surpass the other models and mark the top results at 92.2349\% F1 score. 

Another noteworthy observation in our study was that cased models outperformed all the respective uncased models. 
Even though the xlm-roberta-base was the least performer in discriminative models, the performance gap to the best performer is only 2.3789\% which shows the competitiveness of the transformers in token level classification tasks.

Even though scibert models are specifically trained on scientific corpus, these models were not able to outperform the bert multilingual model, which shows that the general knowledge could play a significant role in metaphoric identification task. 

While ChatGPT seems very good at handling general text, it does not perform well in metaphoric names identification in flower and plant names. Given that we cannot further fine-tune the GPT model with our corpus, the ChatGPT is struggling to identify and generate text with metaphoric flower and plant names. Another important observation was, ChatGPT was not producing consistent results because we could observe different results for the same sentence if we retrieve twice. This shows that ChatGPT is uncertain about its answers on metaphoric flower and plant names, maybe with GPT-4 it may have a better understanding with more data. We leave it for future work. 

\subsection{Spanish} 
Table \ref{tab:spanish_results} shows the results on Spanish data in zero-shot configuration on English data. We note that in all models, learning from English data has lead to decent results on Spanish metaphoric flower and plant names identification. Interestingly, bert-base-multilingual-cased model performs better in both languages marking over 52\% F1 score on Spanish. 
It was noted that there is a significant difference between English and Spanish results, as expected because the English models were fine-tuned on English metaphoric data, but we were not able to do that in Spanish due to lack of resources. 

ChatGPT has kept similar performance for Spanish recording over 51\% F1 score. This is very close value to the best discriminative model but could not outperform bert-base-multilingual-cased model. Unlike ChatGPT, since discriminative models are able to fine-tune, we conjecture that their performance could be boosted with a fine-tuning step with more data.  

\section{Conclusions}
\label{sec:conclusions}
The detection of metaphorical terms is an important research area for many NLP applications. Detecting metaphor-based terms of flowers and plants may give birth to different multidisciplinary research and applications. On the one hand, it may help in overcoming the so-called plant awareness disparity or plant blindness \cite{parsley2020plant} as the metaphoric factor would help in remembering the names of flowers and plants and their aspect. It may also give insightful information to Cognitive Studies towards understanding phenomena such as metaphor and metonymy, and even towards a more comprehensive understanding of conceptual complexes \cite{ruiz2017conceptual}. This may be carried out by comprehending the associations between metaphoric names and the image of the flower and plant representing them, and how the resemblance of images or the metonymic aspect is conceptualised through the coinage of terms. On the other hand, this information is also helpful for the studies of representation of abstract phenomena in art and its comprehension across languages. The automatic extraction of those terms is a step towards achieving more comprehensive and accurate results. In addition, this may help rendering texts more accessible to people with ASD. At the same time, these types of studies may also help in the development of software or mobile applications to be used by both laypersons and professionals.

In conclusion, we show that the state-of-the-art transformers are well capable of performing excellently in identifying metaphoric flower and plant names. 

\section{Acknowledgements}
Part of this research was carried within the framework of the projects the projects PID2020-118369GB-I00 and A-HUM-600-UGR20, funded by the Spanish Ministry of Science and Innovation and the Regional Government of Andalusia. Funding was also provided by an FPU grant (FPU18/05327) given by the Spanish Ministry of Education. We also want to thank Elvira Cámara Aguilera for her help in the annotation process.

\bibliographystyle{fullname}
\bibliography{EjemploARTsepln}

\appendix
\section{Appendix 1: English And Spanish Corpus}
List of references used to compile the corpus in English and Spanish.

\begin{table*}
    \centering
    \scalebox{0.8}{
    \begin{tabular}{|c|}
    \hline
         References \\
         \hline
        \makecell{Brickell, Christopher. Encyclopedia of Plants and \\ Flowers. New York, Dorling Kindersley, 2012} \\
        \hline
        \makecell{Vigneron, Jean Pol, et al. “Optical Structure and \\ Function of the White Filamentary Hair Covering the Edelweiss \\ Bracts.” Physical Review E, vol. 71, no. 1, 19 Jan. 2005} \\
        \hline
        \makecell{Maghiar, Lăcrămioara M., et al. “Integrating Demography \\ and Distribution Modeling for the Iconic Leontopodium Alpinum \\ Colm. In the Romanian Carpathians.” Ecology and Evolution, vol. 11, \\ no. 18, 25 Aug. 2021, pp. 12322–12334} \\
        \hline
        \makecell{Blanco-Pastor, J. L., et al. “Past and Future Demographic \\ Dynamics of Alpine Species: Limited Genetic Consequences despite \\ Dramatic Range Contraction in a Plant from the Spanish Sierra \\ Nevada.” Molecular Ecology, vol. 22, no. 16, \\ 12 July 2013, pp. 4177–4195} \\
        \hline
        \makecell{Ni, Lianghong, et al. “Migration Patterns of Gentiana \\ Crassicaulis, an Alpine Gentian Endemic to the Himalaya–Hengduan \\ Mountains.” Ecology and Evolution, vol. 12, no. 3, Mar. 2022} \\
        \hline
        \makecell{Pink, Alfred. Dictionary of Flowers and Plants for \\ Gardening. Teresa Thomas Bohannon, 2008} \\
         \hline
    \end{tabular}
    }
    \caption{English Corpora}
    \label{tab:english_corpus_table}
\end{table*}

\begin{table*}
    \centering
     \scalebox{0.8}{
    \begin{tabular}{|c|}
    \hline
         Reference \\
         \hline
          \makecell{Gómez García, Daniel. Flora y Vegetación de La  \\ Jacetania. De la Naturaleza. Diputación General de Aragón, 2004} \\
          \hline
          \makecell{M.  López Guadalupe, et al. “Comunidades, Hábitat Y Tipos de \\ Suelos Sobre Los Que Se Desarrolla La Manzanilla de Sierra Nevada.” \\ Ars Pharmaceutica (Internet), vol. 26, no. 4, 1985, pp. 255–263} \\
          \hline
          \makecell{Pugnaire, Francisco, et al. “Facilitación de Las Especies \\ Almohadilladas Y Cambio Global En Las Comunidades Alpinas Del Parque \\ Nacional de Sierra Nevada.” Proyectos de Investigación En Parques \\ Nacionales: 2010-2013, by Ministerio de Agricultura, Ministerio de \\ Agricultura, Alimentación y Medio Ambiente. Organismo Autónomo de \\ Parques Nacionales, 2015, pp. 91–104} \\
          \hline
          \makecell{Montserrat, P, and R Balcells. “LA FLORA DEL PIRINEO.” \\ Sinergia (Publicación Paramédica de Sociedad General de Farmacia, SA), \\ vol. 14, no. 6675, 1960} \\
          \hline
          \makecell{Paúl Gonzáles, et al. Las Plantas Comunes Del Bosque Seco \\ Del Marañón: Biodiversidad Para Las Comunidades Locales | Bosques \\ Andinos. Bosques Andinos, Lima- Perú, Biblioteca Nacional del Perú, \\ 2020} \\
          \hline
          \makecell{Blanca, Gabriel. Flora Amenazada Endémica de Sierra Nevada. \\  Granada, Universidad de Granada, 2001}\\
          \hline
         \makecell{Peñas, Julio, and Juan Lorite. “Biología de La Conservación \\ de Plantas En Sierra Nevada. Principios Y Retos Para Su Preservación \\  - Universidad de Granada.” Editorial.ugr.es, Universidad de \\ Granada, 2019} \\ 
         \hline
         \makecell{Ministerio de Salud y Protección Social de Colombia. \\ “Vademecum Colombiano de Plantas Medicinales.” Orasconhu.org, \\ Ministerio de Salud y Protección Social de Colombia, 2008} \\
         \hline
         \makecell{Sánchez de Lorenzo Cáceres, Jose Manuel. Los Árboles En \\ España: Manual de Identificación. Ediciones Mundi-Prensa, 1999} \\
         \hline
    \end{tabular}
    }
    \caption{Spanish Corpora}
    \label{tab:spanish_corpus_table}
\end{table*}



\end{document}